\documentclass[11pt,a4paper]{article}
\usepackage{emnlp2021}
\usepackage{times}
\usepackage{latexsym}
\usepackage{xcolor}
\usepackage{graphicx}

\usepackage{microtype}

\usepackage{color}
\usepackage{xspace}
\usepackage{multirow}
\usepackage{subcaption}
\usepackage{float}
\usepackage[skip=1pt]{caption}

\usepackage[symbol]{footmisc}

\newcommand\blfootnote[1]{%
  \begingroup
  \renewcommand\thefootnote{}\footnote{#1}%
  \addtocounter{footnote}{-1}%
  \endgroup
}

\title{\textsc{MindCraft}:\\Theory of Mind Modeling for Situated Dialogue in Collaborative Tasks}

\author{
    Cristian-Paul Bara$^*$ \\
    University of Michigan \\
    \texttt{cpbara@umich.edu} \\
    \And
    Sky CH-Wang$^{*\dagger}$ \\
    Columbia University \\
    \texttt{skywang@cs.columbia.edu}\\ 
    \And
    Joyce Chai \\
    University of Michigan \\
    \texttt{chaijy@umich.edu} \\
    }

\date{}

\begin{document}
\maketitle
\begin{abstract}

An ideal integration of autonomous agents in a human world implies that they are able to collaborate on human terms. In particular, theory of mind plays an important role in maintaining common ground during human collaboration and communication. To enable theory of mind modeling in situated interactions, we introduce a fine-grained  dataset of collaborative tasks performed by pairs of human subjects in the 3D virtual blocks world of Minecraft. It provides information that captures partners' beliefs of the world and of each other as an interaction unfolds, bringing abundant opportunities to study human collaborative behaviors in situated language communication. As a first step towards our goal of developing embodied AI agents able to infer belief states of collaborative partners in situ, we build and present results on computational models for several theory of mind tasks. 

\end{abstract}
\blfootnote{$^*$Equal Contribution.}
\blfootnote{$^\dagger$Work performed while the author was an undergraduate research assistant at the University of Michigan.}

\vspace{-8pt}
\section{Introduction}
\vspace{-4pt}
Creating embodied, situated agents able to move in, communicate naturally about, and collaborate on human terms in the physical world has been a persisting goal in artificial intelligence \cite{winograd1972understanding}. 
During communication in such a setting, agents not only need to ground entities in language to that of the physical world; efficient and accurate human-agent collaboration further requires agents to reason about the progress of the task at hand and to plan and execute a series of collaborative steps, whilst maintaining common ground \cite{clark1996using} with collaboration partners, in order to achieve a certain goal. 

Despite recent advances, we are still far away from fully enabling these desired agent behaviors. One key challenge is in an agent's ability to establish and maintain common ground in tandem with human partners, especially in a setting where beliefs about the world and of each other may change on the fly \cite{popat2005creating,powers2005common}. It is important to understand how changes in a dynamic, physical world affect agents' beliefs of each other—i.e., theory of mind \cite{premack1978does}—and how such beliefs influence teamwork and communication in collaborative tasks.
As a first step to address this question, this paper explores theory of mind modeling \cite{chandrasekaran2017takes,rabinowitz2018machine,jara2019theory} in situated language communication \cite{iwahashi2009robots,mcguire2002multi} for collaborative tasks within the 3D virtual blocks world of Minecraft.
Through a novel experimental setup, we collect a situated dialogue dataset that demonstrates how collaborative partners with a set of asymmetric knowledge and skills are able to collaborate to achieve joint goals, and how, in particular, their beliefs of each other evolve and converge over time. Based on this dataset, we further build several baseline computational models to explicitly predict key elements of a collaboration partner's mental state from the viewpoint of an agent as a task unfolds. Our empirical results demonstrate that while language is certainly important in this inference, the shared physical environment and the perceived activities play a greater role in shaping a partner's understanding of each other in order to come to a common ground.


%
The contributions of this work are threefold. First, we introduce \textsc{MindCraft}, a task in which pairs of users collaboratively work to create novel materials by combining blocks in the 3D virtual world of Minecraft, with the ultimate objective of creating a final, goal material. Unlike prior work in situated collaborative tasks \cite{liu13, bisk2018learning, suhr2019executing}, a key focus of our work is to facilitate theory of mind modeling—the ability to attribute mental states, both of one's own and that of others—an important but not yet well-studied topic in situated collaborative interactions.
Within designed collaborative tasks, we have users record their beliefs about the state of the game, and of each other, at periodic intervals. Our data captures an evolution of the states of mind of our participants that are true representations of their beliefs—not simply proxies for the true sequence of events in a collaborative session. This explicit modeling of theory of mind sheds light on how partners strive to align their mental models in order to achieve common ground during collaboration. 

Second, departing from previous Leader-Follower setups (where one partner explicitly leads and gives instructions to the other, the follower, who tries to execute said instructions)~\cite{suhr2019executing}, we focus on a setting where partners each have asymmetric knowledge \cite{bortolaso2019enhancing} and skill-sets towards completing a joint goal. In order to effectively complete the given tasks, partners need to \textit{negotiate their own plans of action} by taking into account what they currently know and don't know about their partner, and of their common understanding of the task at hand. Our novel, more relaxed setup provides support for greater diversity in modes of collaboration that are more representative of that in the real world.

Third, we introduce a set of baseline computational models to infer fellow player mental states in situ, as a collaborative agent would, and highlight some further challenges present in moving towards building fully realistic agents able to reason about human mental states in situated environments. 

Our platform, data, and models are made available$^\ddagger$ \blfootnote{$^\ddagger$\url{https://github.com/sled-group/MindCraft}} and will facilitate future work on physical agents that can effectively collaborate with humans through situated dialogue. 
\section{Related Work}

Our work builds upon existing efforts within collaborative dialogue in understanding the nuances of human collaboration and in towards building computational agents that can engage in language communication and collaborative tasks with humans in a physical environment. 

Situated and task-oriented natural language interactions \cite{iwahashi2009robots,zarriess2016pentoref} have been studied in a variety of environments, including in custom 2D worlds \cite{liu12sigdial, udagawa2020annotated,udagawa2019natural}, in the physical world with human-robot interactions \cite{mcguire2002multi,chai2014collaborative, chai2018language, thomason2020jointly}, and in various 3D virtual worlds \cite{bisk2018learning,suhr2019executing}. Most closely, our environment builds upon recent work by \citet{narayan2019collaborative} and \citet{jayannavar2020learning}, whereby computational models of user dialogue prediction and user next-action prediction are investigated in the setting of a collaborative dialogue task within the 3D virtual blocks world of Minecraft. 
{However, to our knowledge, none of these previous works explicitly model theory of mind for dialogue agents.}

Theory of mind as a subject, especially in computation \cite{Laird_Lebiere_Rosenbloom_2017}, has gained increased attention in areas including agent-agent reinforcement learning \cite{rabinowitz2018machine}, dialogue systems \cite{qiu2021towards}, human-computer interaction \cite{wangCHI21tom}, agent-agent collaborative dialogue \cite{roman2020rmm}, and explainable AI \cite{akula2021cxtom}. Worthy of note is the type of mental state recording we employ: specifically, we ask players to record their own mental states during interaction. Unlike prior work that has largely utilized external annotators for post-hoc mental state attribution
, we expand on \citet{eicher2017toward} and \citet{wangCHI21tom} by specifically bringing user self-reported mental states from that of only the linguistic domain to multimodal situated dialogue. Specifically, the novelty in our work exists in studying and bringing explicit theory of mind modeling to 3D situated collaborative interactions.






\section{Experimental System and Data Collection}
\label{sec:dataset}

\begin{figure*}[t]
    \centering
    \includegraphics[width=1.0\textwidth]{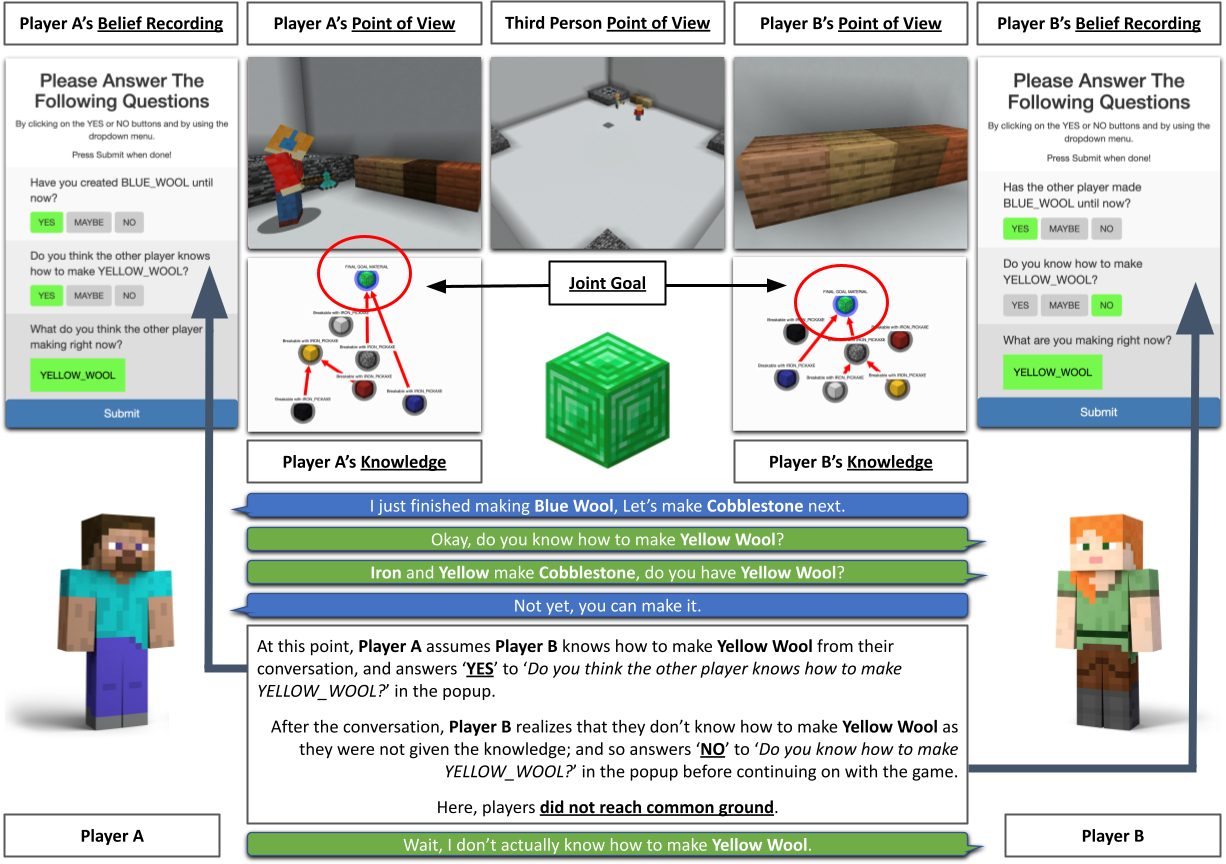}
    \caption{Diagram of a sample interaction in \textsc{MindCraft}. Two players are tasked to complete a common goal within the game environment of Minecraft; players communicate using in-game chat, are provided partial views of the plan needed to create the goal material, and are periodically asked paired questions to probe into their mental states. Additionally, we record first-person viewpoint videos of the two players' points of view (POV) as well as a third-person POV from the shared game environment.}
    \label{fig:data_cartoon}
\end{figure*}


We consider a scenario whereby two agents, situated in the same environment and able to perform actions simultaneously, collaborate to complete a shared goal. Here, unlike traditional Leader-Follower setups, \textit{both} agents have \textit{asymmetric information} on the steps needed to complete the target task. In addition, in certain iterations, agents have \textit{asymmetric skill-sets} as well: certain steps may only be completed with specific skills, and an agent may not be able to complete the target task by themselves, even with complete knowledge. Agents are provided a text channel to communicate in natural language, where they are able to share knowledge and negotiate on the actions to be performed by each agent in the process of completing the target task. Agents can also directly perform actions in the environment based on their own, albeit partial plan, and on their current understanding of the game state. We study this scenario in a modified blocks world environment of Minecraft with our custom game, \textsc{MindCraft}. 
Figure~\ref{fig:data_cartoon} gives an overview of our experimental system and setup. 


\subsection{\textsc{MindCraft}}
The goal of \textsc{MindCraft} is to create a specific goal material that is randomly generated for each game. A set of material blocks are spawned in the environment when agents enter, serving as the starting set of materials of the game. Agents have two \textit{macro}-actions to create new materials:
\begin{itemize}
\vspace{-8pt}
\item \texttt{Mining}, where agents hit a specific block to create a new block type. This may be repeated ad infinitum to create unlimited new blocks of that type. 
\vspace{-8pt}
\item  \texttt{Combining}, where agents stack two specific blocks on top of each other, consuming them and creating a single block of a new type in their place. 
\vspace{-8pt}
\end{itemize}

Note that \textit{macro}-actions themselves are composed of many fine-grained, \textit{atomic} actions that players may perform in-game, such as moving around, breaking blocks, chatting, jumping, and more—which utilize the full capability of the Minecraft game environment.



\subsection{Modeling Agent Knowledge and Skills}

In the real world, agents that engage in collaborative tasks may each have partial knowledge and incomplete skill sets. We are particularly interested in how these agents collaborate and negotiate with each other to come to a shared plan in order to achieve a joint goal. To this end, we explicitly model agents' knowledge and skills in \textsc{MindCraft} tasks.

{\bf Knowledge.} Each player is given a knowledge graph—the recipe—as shown in Figure~\ref{fig:data_cartoon}. Recipes given to players specify a joint goal and a {\em partial} set of macro-actions needed to take place toward completing the goal. For example, Player A, from the initial recipe, knows how to create \texttt{Yellow Wool}, but they do not know how it would contribute to making the goal material, \texttt{Emerald Block}. On the other hand, Player B, while they do not initially know how to create \texttt{Yellow Wool}, is given the knowledge that doing so would lead to creating \texttt{Cobblestone}, then used to make the goal material.

{\bf Skill-Sets.} In order to stack blocks together, agents must be able to physically move blocks around in the virtual environment. This is achieved by hitting blocks with specific tools; randomly generated constraints exist in each game that specify which tools are able to interact with which blocks. Given randomly to agents at the start of each game, these tools effectively set constraints on which agents possess the necessary skills to interact with certain block types. 

Combined, this asymmetry in both knowledge (recipes) and skill-sets (tools) motivates communication, as each individual agent does not (1) know how to create the goal material (2) nor do they have the skill-set to do so on their own. Furthermore, as both agents are situated within the environment, both have only partial observability of the game state, limited by their first-person field of view in-game. Players need to collaborate and communicate with each other to achieve the joint goal. 

\subsection{Belief Modeling and Common Ground}
We facilitate theory of mind studies by asking players to record their beliefs about the progress of the current game, and of each other, at periodic intervals. As shown in Figure~\ref{fig:data_cartoon}, each player is asked three types of questions:
\begin{itemize}
    \item {\bf Completed Task Status.} This asks if a specific material has been created, by themselves or by the other player, since the start of the game, probing into the player's beliefs about the current state of the game as influenced by either themselves or their collaboration partner. For example, as shown in Figure~\ref{fig:data_cartoon}, Player B is prompted with the question ``Has the other player made \texttt{Blue Wool} until now?''
    \vspace{-8pt}
    \item {\bf Player Knowledge.} This asks if the player \textit{knows} how to create a specific material, or if they \textit{believe that their partner possesses the knowledge} to create it. This probes into a player's current knowledge of their own and of their partner's current knowledge, as influenced by the initial knowledge they were provided and that which has been gained, via communication with their partner, since the start of the game. In this example, Player B is given the question ``Do you know how to make \texttt{Yellow Wool}?'' 
    \vspace{-8pt}
    \item {\bf Player Current Task.} This asks players what they believe they themselves are making, or believe their partner to be making, at \textit{current} time. For example, Player A is given the question ``What do you think the other player is making right now?''
\end{itemize}
\textbf{Question Pairing.} The three questions received by players are paired by type; i.e. if one player is asked a question of their own beliefs, the other player is asked the same question on what they believe their partner's beliefs to be. In the example given, when Player B is prompted with the question ``Has the \textit{other player} made \texttt{Blue Wool} until now?'', at the \textit{same time}, Player A is prompted with the question ``Have \textit{you} created \texttt{Blue Wool}?'' The game is paused when players record their answers to the set of questions and resumed when both players have completed their answers.

By explicitly soliciting players' states of mind during collaboration, we are able to define a quantitative measure of common ground: specifically, we consider common ground to be instances of answer agreement among pairs of players to a given question.

\subsection{Data Collection}
With the experimental setup described above, we collected a dataset totaling 100 games. Pairs of players participated in the experiments through a remote video conference, where they were instructed to access a custom Minecraft server using a game client, as well as a web page interface that we provided, used to display recipe information and to collect player beliefs with periodic popups, once every 75 seconds. Pop-ups ask three questions at once\textemdash one of each type\textemdash the content of each being paired to the corresponding question asked to the other player. During games, players are only able to communicate with each other using in-game chat, and each pair of players played at most 5 games.


From these games, we log their timestamped dialogue utterances via in-game chat, their questions and answers to the periodic popups for belief recordings, an internal game log that stores the entire game state, and three sets of video recordings, representing each player's first-person point of view and a third-person point of view at a high vantage point with a clear view of the entire game. 

In our dataset, there is an average of 20.5 dialogue exchanges per game, for a total of 2091 exchanges. Games last between 1 minute and 22 seconds to 27 minutes and 26 seconds, with the average game lasting 7 minutes and 23 seconds. A total of 12 hours, 18 minutes, and 33 seconds of in-game interaction was recorded. On average, 4 popup question pairs appear each game. Between 5 and 10 objects are used in a game, and between 7 and 11 macro steps are necessary in each game to achieve the goal.

\section{Findings and Observations}
\label{sec:findings}

We further perform an analysis of our dataset to gain an understanding of collaborative behaviors between players both in their reasoning and in their alignment of mental models.

\subsection{The Role of Asymmetry in Knowledge and Skill-Sets}

\begin{table*}[h!]
    \centering
    \resizebox{\textwidth}{!}{
    \begin{tabular}{cc|rrr|rrr|rrr}
        Skills    & Knowledge & \multicolumn{3}{c|}{Dialogue Exchanges} & \multicolumn{3}{c|}{Duration (minutes)} & \multicolumn{3}{c}{Agreement} \\
                  &           & \multicolumn{3}{c|}{} & \multicolumn{3}{c|}{} & Completed  & Knowledge  & Current \\
                  &           & min & avg(std) & max & min & avg(std) & max & Task && Task \\
                  \hline
                  
   Shared &    Shared &        3 &     10(\phantom{0}7) &        27 &    1:23 &   4:25(2:55) &  11:02  &        0.706 &         0.529 &         0.176 \\
   Shared & Disparate &        2 &     15(11) &        37 &    2:29 &   6:52(4:07) &  13:46   &      0.654 &         0.731 &         0.231 \\
Disparate &    Shared &        5 &     22(16) &        54 &    2:49 &   8:05(5:35) &  19:26    &     0.778 &         0.593 &         0.444 \\
Disparate & Disparate &        6 &     28(22) &        72 &    2:12 &   7:49(5:21) &  18:15     &    0.654 &         0.385 &         0.308 \\

    \end{tabular}}
    \caption{Statistics on games with varying skill and knowledge configurations; minimums (min), averages (avg), maximums (max), and standard deviations (std) for the number of dialogue exchanges and durations of each game configuration are shown, as are player agreements for all three question types.}
    \label{tab:disparity_stats}
\end{table*}

To quantitatively understand how a disparity in skill-sets and knowledge affects player behavior in situated collaborative tasks, we perform an initial pilot study based on four different configurations that vary on whether players share the same, complete plan (i.e. knowledge) and/or the same tools (i.e. skills) necessary to complete the task. In disparate configurations, both players possess disparate, partial plans and/or tools, with partial overlap between players. 
This pilot consists of 32 games to measure key statistics in areas of communication, interaction length, and mutual mental state agreement, with 8 games per configuration. Games were played in sets of four between pairs of players in a round-robin fashion across configurations, mitigating for external factors among pairs such as player game and mutual familiarity.


As shown in Table \ref{tab:disparity_stats}, within our expectation, a disparity in both skill-sets and knowledge causes players to disagree and communicate the most to a statistically significant degree, and a disparity in either produces significantly more dialogue utterances than when both are shared. 
Players in fully disparate games have the lowest agreement in mutual knowledge and task completion.
Despite this, in fully disparate games, a higher level of agreement is present for beliefs of the current tasks being performed by either player (e.g., compared to the shared skills configuration), which we attribute to players needing to ask for help more, thus communicating more and being more aware of each other, in such situations. 

\begin{figure*}[h!]
    \centering
    \begin{subfigure}{0.32\textwidth}
        \includegraphics[width=\textwidth]{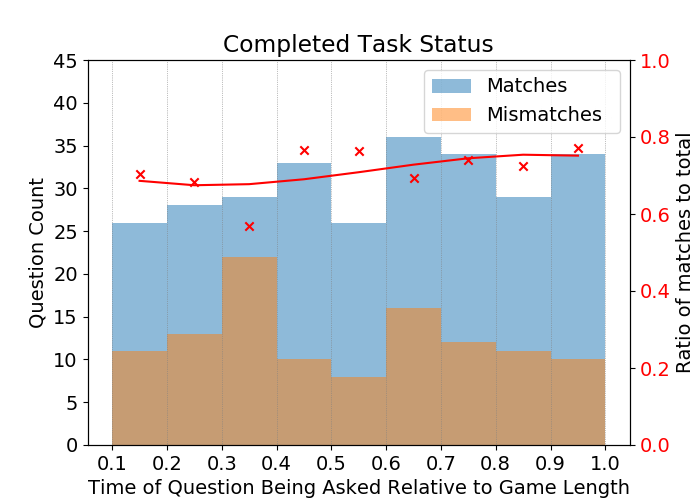}
        \caption{}
        \label{fig:hist_human_label1} 
    \end{subfigure}
    \begin{subfigure}{0.32\textwidth}
        \includegraphics[width=\textwidth]{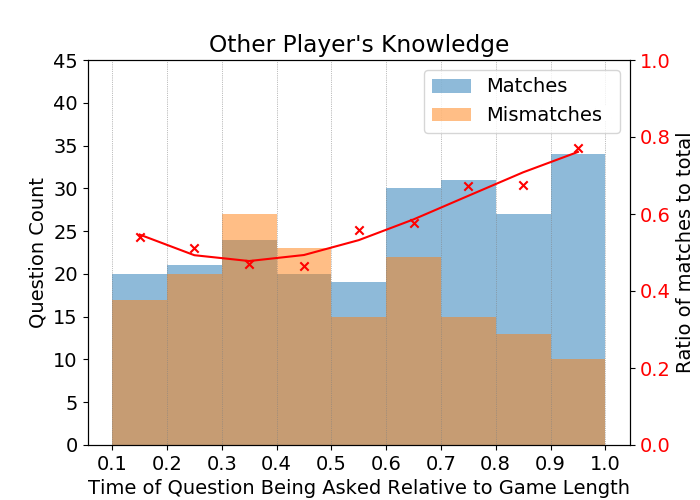}
        \caption{}
        \label{fig:hist_human_label2} 
    \end{subfigure}
    \begin{subfigure}{0.32\textwidth}
        \includegraphics[width=\textwidth]{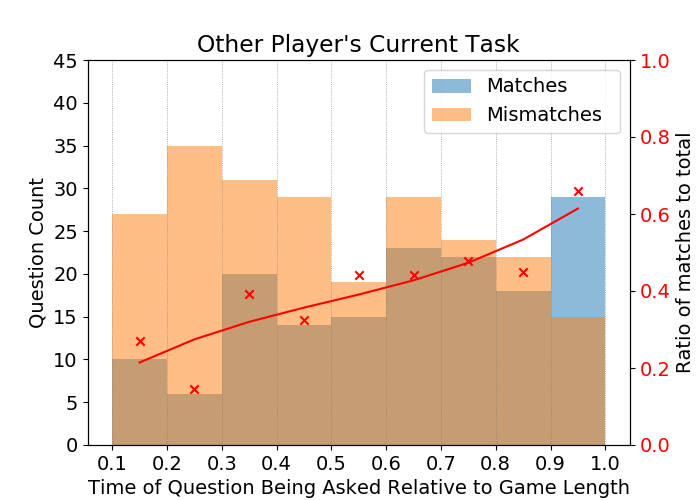}
        \caption{}
        \label{fig:hist_human_label3} 
    \end{subfigure}
    \caption{Histograms of \emph{player} answer matches (agreement, in blue) and mismatches (disagreement, in orange) to question pairs on (a) completed task status, (b) player knowledge, and (c) current task status asked at different relative intervals during games. Red crosses show the ratio of matched answers out of the total questions, and the red line shows the $3^{rd}$ order polynomial fit to the crosses.}
    \label{fig:hist_human_label}
\end{figure*}

\subsection{Evolution of Belief States}

To understand how the interaction discourse shapes partners' beliefs of the tasks and of each other, we take a closer look at three types of beliefs (as reflected by our three types of questions) and examine how they evolve as collaboration and communication unfold. Segmenting individual games into 10\% sections across each game's duration, we examine player agreement and disagreement as games progress. Figure~\ref{fig:hist_human_label} shows the aggregated results from all games for our three types of beliefs.




On average, player agreement on completed task status remains high and relatively constant throughout a game's progression, averaging around 80 percent, as shown in Figure ~\ref{fig:hist_human_label1}. 
%
%
However, as each game progresses, there is a noticeable increase in the agreement among two players in terms of what they believe about the other player's knowledge (Figure~\ref{fig:hist_human_label2}). 
%
%
Similarly, beliefs about what the other player's current task is also increase notably in agreement as each game progresses, averaging around 12 percent at the start, gradually reaching over 60 percent by the end of the game (Figure~\ref{fig:hist_human_label3}).

These results demonstrate that the longer the two players collaborate with each other in a game, the more aligned they become in their beliefs about each other. Furthermore, player understanding of completed tasks can be acquired by direct observations from the environment itself, and it's easier to reach an agreement (i.e., joint understanding or common ground) here than an understanding of a partner's mental states.


\begin{figure*}[h!]
    \centering
    \begin{subfigure}{0.32\textwidth}
        \includegraphics[width=\textwidth]{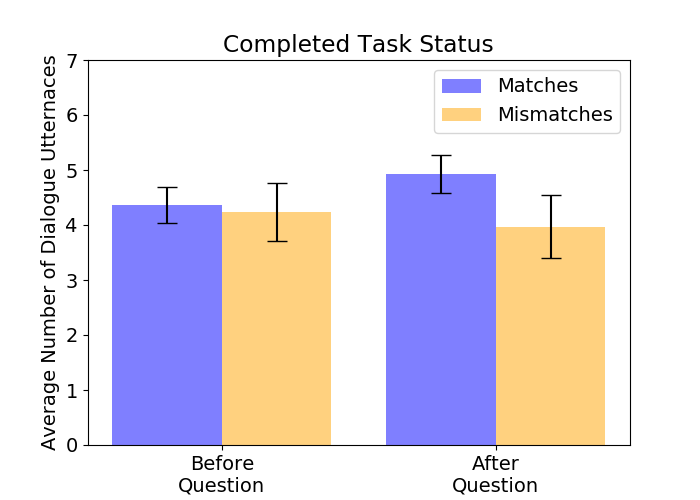}
        \caption{}
        \label{fig:avg_dlg1} 
    \end{subfigure}
    \begin{subfigure}{0.32\textwidth}
        \includegraphics[width=\textwidth]{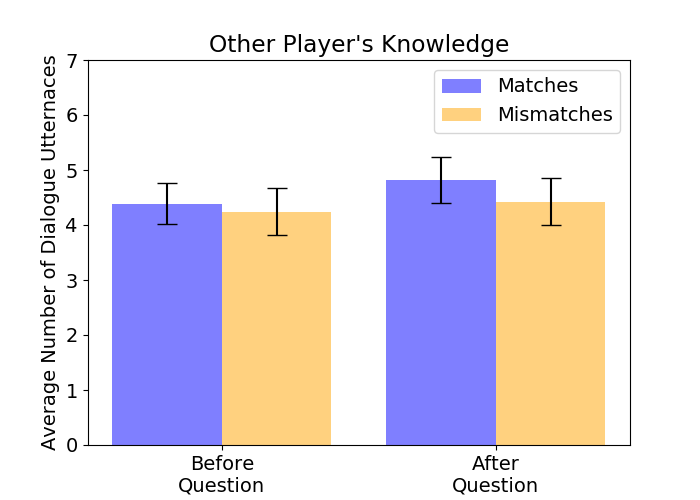}
        \caption{}
        \label{fig:avg_dlg2} 
    \end{subfigure}
    \begin{subfigure}{0.32\textwidth}
        \includegraphics[width=\textwidth]{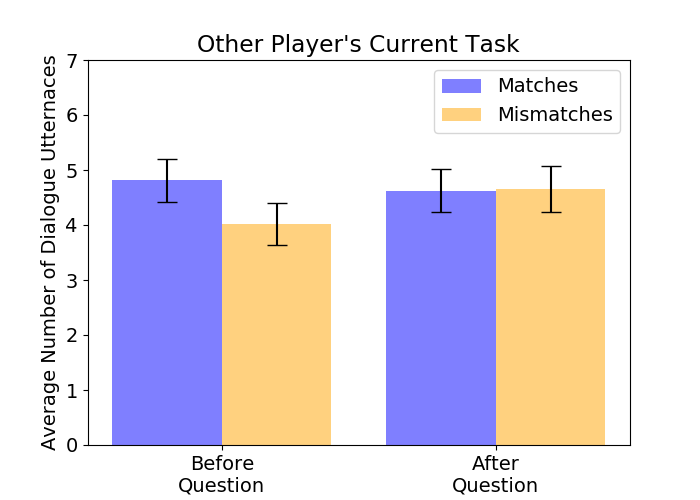}
        \caption{}
        \label{fig:avg_dlg3} 
    \end{subfigure}
    \caption{Average number of dialogue exchanges before and after a question on (a) completed task status, (b) player knowledge, and (c) current task status was asked for cases of player agreement (match) and disagreement (mismatch). 
    }
    \vspace{-1em}
    \label{fig:avg_dlg}
\end{figure*}

\subsection{Dialogue Behavior}

To better understand how agreement or disagreement in players' mutual beliefs affect dialogue behavior, we conduct a further analysis by examining dialogue utterances in a fixed time window of 75 seconds before and after a question is asked for each question type, separating instances of agreement and disagreement. Figure~\ref{fig:avg_dlg} shows the average number of dialogue exchanges across all games in this stratification.


For beliefs about the status of a completed task (Figure \ref{fig:avg_dlg1}), we observe no difference in dialogue exchanges before the question is posed between instances of agreement and disagreement in beliefs; interestingly, however, immediately following a given question, a significant difference becomes apparent in the number of dialogue exchanges. When there is \textit{agreement} between players about the state of tasks, we observe that they, on average, tend to communicate \textit{more} to continue on the course to further elaborate on their plan. 


On the other hand, for beliefs of partner knowledge, we do not observe a change in behavior before or after a question is asked (Figure \ref{fig:avg_dlg2}), and, for beliefs that involve a partner's current task (Figure \ref{fig:avg_dlg3}), interesting of note is that the average number of dialogue exchanges leading to disagreement was significantly less than that which led to agreement. This highlights a potential reason why disagreement occurred: less communication. We observe that communication is especially important for players to infer what tasks their partner is currently working on, as it's difficult to know the current goal of a partner by only observing their partial actions without communicating about it, due to their own incomplete plan. 

\section{Computational Models for Inferring Belief States}

\begin{figure*}[t]
    \centering
    \includegraphics[width=\textwidth]{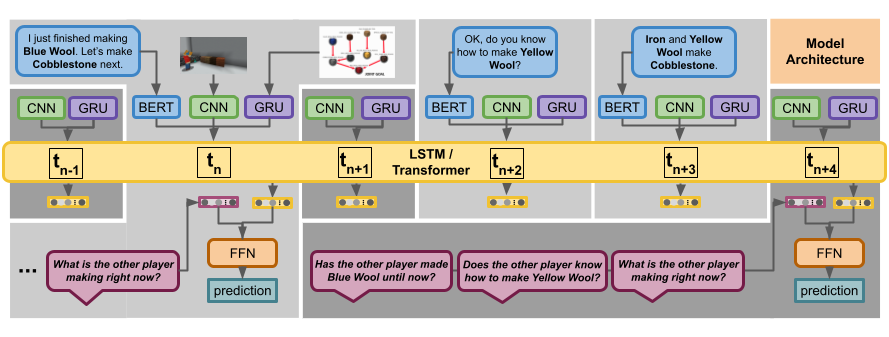}
    \caption{Model Architecture. Each time step is sampled once every second, where we take one frame for each time step. 
    All time steps, shown in light and dark gray boxes, have video frames associated with them, but not all have dialogue utterances (e.g. in the dark gray time-steps above) or questions, as players might not have chatted with each other or been prompted with questions at every time step. Player dialogue utterances are shown in blue; prompted questions are in purple.}
    \label{fig:model_architecture}
\end{figure*}

Based on our dataset, a variety of computational problems can be formulated, developed, and evaluated. In this section, we focus on one key problem—predicting player belief states of the task and of a collaborative partner in situ. As a first step, we implement a straightforward model that, from a player's perspective, predicts the state of the task as well as the mental states of a partner at any given time based on historical observations of a rich discourse of dialogue and perceived actions in the shared environment. 

Figure~\ref{fig:model_architecture} shows the overall architecture of our model. Our dataset is comprised of two time-series-based modalities: (1) a video stream coming from either player's first-person POV, and (2) dialogue exchanges. We implement a forward sequence-to-sequence model, such that inferences at any given time are only able to process inputs that have occurred before it.




\vspace{5pt}
\noindent
\textbf{Plan Processing.} Recall that each player is provided a partial view of the complete plan. Here, each plan is stored as a list of tuples, representing each material present in the plan, associated with the materials needed to make it and the tool needed to interact with it. Represented naturally as a graph, the list of nodes is given as input to a GRU \cite{chung2014empirical} for encoding. In tasks that involve predicting a partner's mental state in situ, only the partial plan associated with the player (not the partner) is used.

\vspace{5pt}
\noindent
\textbf{Visual and Dialogue Processing.} We encode video frames with a Convolutional Neural Network and encode dialogue utterances with  \texttt{bert-large-uncased} \cite{Devlin2019BERTPO}. As dialogue exchanges are a sparse input, a zero-vector is used when there is no associated dialogue utterance at a particular time point. 


\vspace{5pt}
\noindent
\textbf{Time Series Processing.} We use either an LSTM \cite{hochreiter1997long} or a Transformer network \cite{vaswani2017attention}, masked such that it only attends to the past, and feed a sequence of visual frame, plan, and dialogue embeddings as aforementioned to produce a latent representation of game interactions for every step.

\vspace{5pt}
\noindent
\textbf{Learning and Inference. } 
Questions, together with each question's associated game embeddings (i.e., dialogue utterance embeddings, visual frame embeddings, and the agent's own partial plan embedding) at corresponding time steps pass through a Feed-Forward Network to make predictions of their answers. Ground truth answers to questions and cross-entropy loss are used for model training. The same overall architecture is used for all question types; the only difference between them is the space of their output predictions.

\begin{figure*}[h!]
    \centering
    \begin{subfigure}{0.32\textwidth}
        \includegraphics[width=\textwidth]{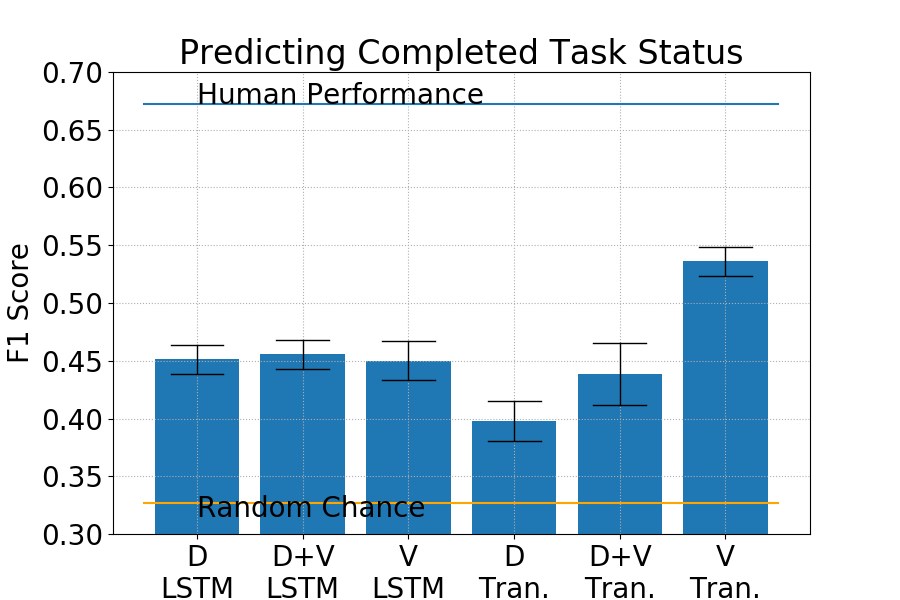}
        \caption{}
        \label{fig:exp_results1} 
    \end{subfigure}
    \begin{subfigure}{0.32\textwidth}
        \includegraphics[width=\textwidth]{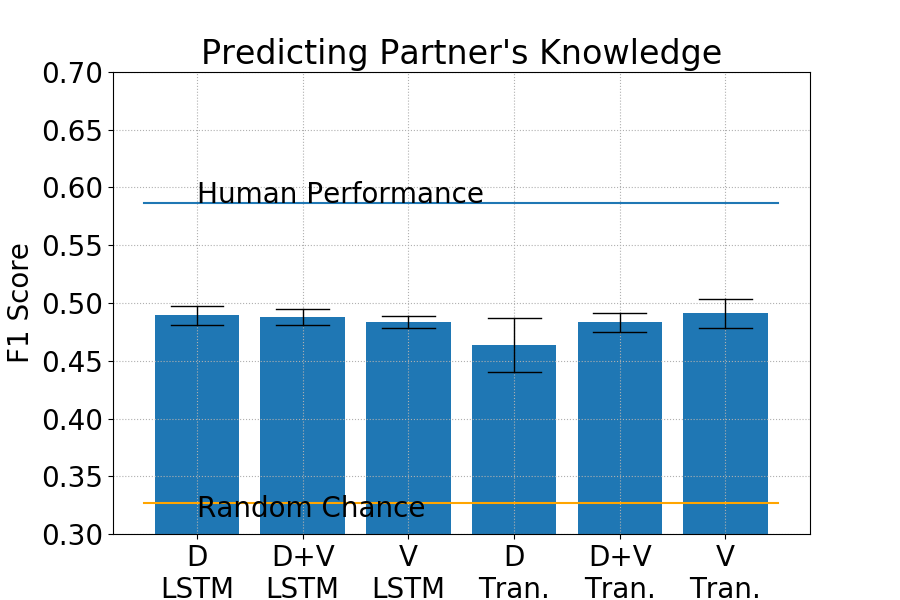}
        \caption{}
        \label{fig:exp_results2} 
    \end{subfigure}
    \begin{subfigure}{0.32\textwidth}
        \includegraphics[width=\textwidth]{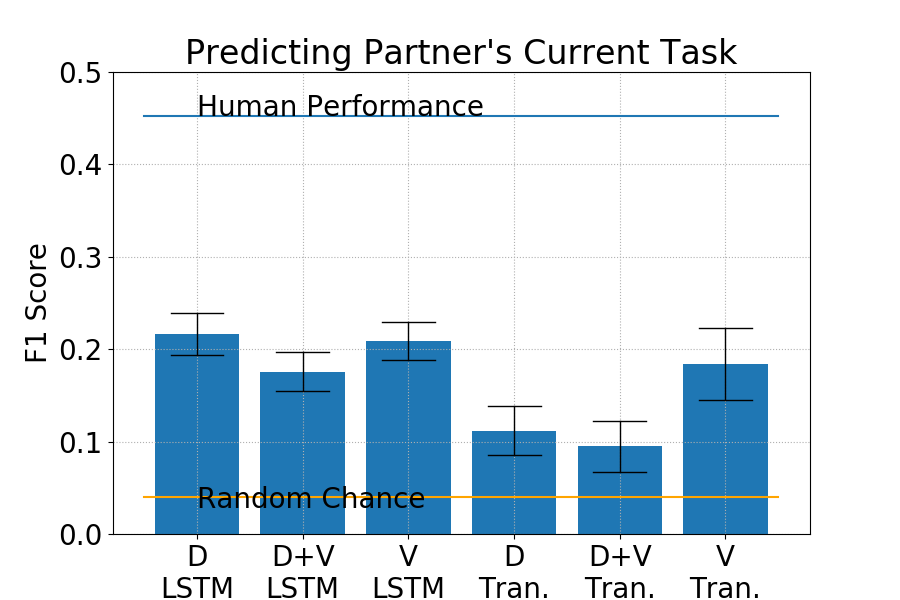}
        \caption{}
        \label{fig:exp_results3} 
    \end{subfigure}
    \caption{Model F1 scores on predicting player belief states. Human performance and random chance performance are marked by the blue and orange horizontal lines, respectively. Detailed results are given in Appendix Table \ref{tab:detail}.}
    \vspace{-1em}
    \label{fig:exp_results}
\end{figure*}
\vspace{-4pt}

\section{Evaluation of Belief State Inference}



We randomly partition our dataset into 60\%/20\%/20\% training, validation, and testing splits with the condition that all three partitions have a similar distribution of game lengths. To achieve testing in situ, we replace one of the two players with our model. At every point where a question is prompted, our model is used to provide an answer about the other player's belief state through inference, using the self-reported belief state of the other player as ground truth for evaluation. We present the multi-class average F1 score weighted by the number of instances in each class (accounting for class imbalances) in our results. We perform our experiments by varying the following configurations:
\begin{itemize}
\vspace{-5pt}
    \item Neural architecture: LSTM or Transformer, with the rationale that they have different abilities in capturing long-distance dialogue history.
    \vspace{-8pt}
    \item Input: dialogue exchanges only (\texttt{D}), first-person POV video stream only (\texttt{V}), and both (\texttt{V+D}), with the intent to understand the role of both language communication and visually perceived activities in the environment towards the task of mental state inference. 
\end{itemize}


\subsection{Performance in Situ}

\textbf{Inferring Player Beliefs of Completed Tasks.} 
Here, we predict player beliefs on the subject of task completion: whether a designated sub-task has been completed by their partner, specifically in response to questions such as ``Has the other player made an Emerald Block until now?''. Participant answers can be one of \texttt{Yes}, \texttt{Maybe}, or \texttt{No}.
This experiment aims to gauge an agent's ability to keep track of the two player's progress towards their goal based on its own knowledge (i.e., the partial plan available to the agent) and the shared interaction history. 
As shown in Figure \ref{fig:exp_results1}, 
we find that the best performing configuration is the Transformer-based model that uses only the video modality.  
This result seems to suggest that \emph{Seeing is believing}; in situated communication, as partners are co-present in a shared environment, they can observe each other's activities and the resulting world state after participant actions to reason about completed tasks—collaborators don't need to use language to communicate about what has already been accomplished. Furthermore, as a large time period may exist between sub-tasks that have been completed and their associated belief question prompts, the Transformer-based model with video inputs only is able to significantly outperform LSTM-based models which may be unable to capture such a time dependency.

\vspace{3pt}
\noindent\textbf{Inferring Player Beliefs of Partner Knowledge.} 
Here, we predict player beliefs of the knowledge possessed by their partners to achieve designated sub-goals, specifically in response to questions such as ``Does the other player know how to make an Emerald Block?''. Participant answers can be one of \texttt{Yes}, \texttt{Maybe}, or \texttt{No}.
Our results in Figure \ref{fig:exp_results2} show that different model configurations result in similar performance, as players are able to explicitly ask questions about each other's knowledge in dialogue exchanges (Figure~\ref{fig:data_cartoon}) in addition to making their own observations from the environment and inferring directly from the plan they were given.

\begin{figure*}[t]
    \centering
    \begin{subfigure}{0.32\textwidth}
        \includegraphics[width=\textwidth]{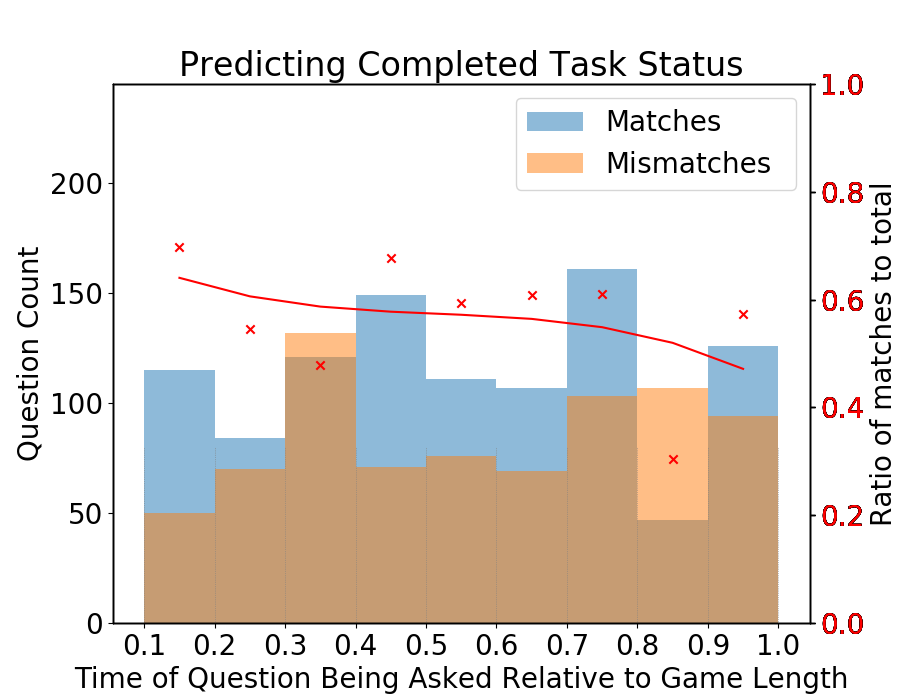}
        \caption{}
        \label{fig:hist_results1} 
    \end{subfigure}
    \begin{subfigure}{0.32\textwidth}
        \includegraphics[width=\textwidth]{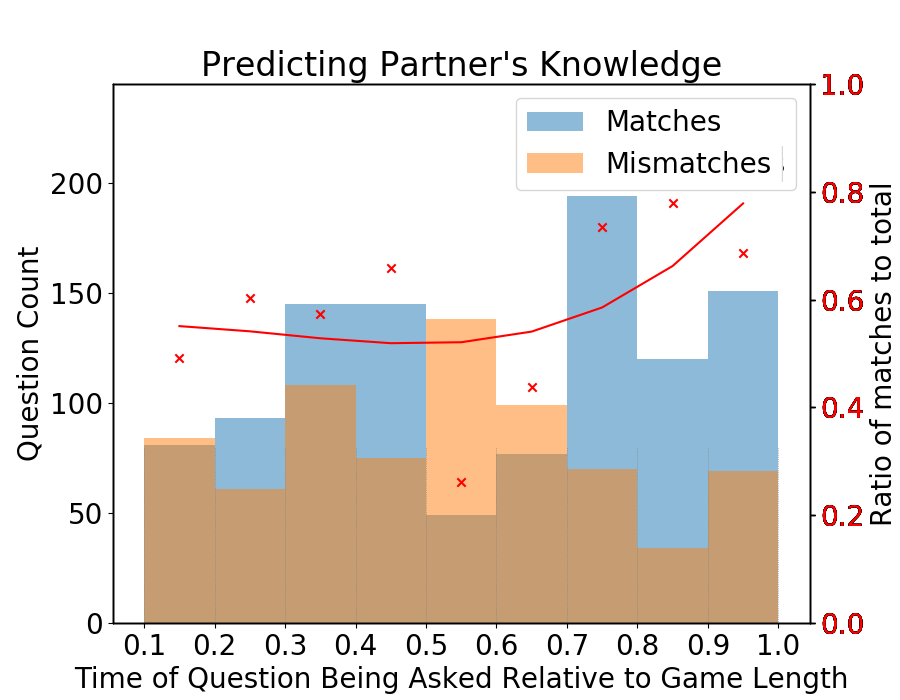}
        \caption{}
        \label{fig:hist_results2} 
    \end{subfigure}
    \begin{subfigure}{0.32\textwidth}
        \includegraphics[width=\textwidth]{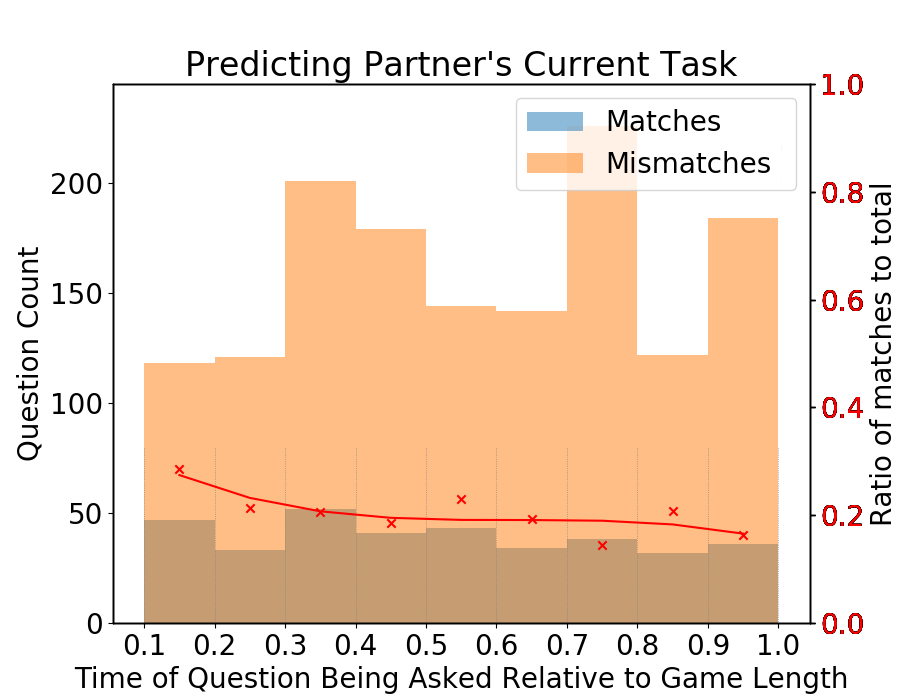}
        \caption{}
        \label{fig:hist_results3} 
    \end{subfigure}
    \caption{Histograms of test-set \emph{model-predicted} answer matches (agreement, in blue) and mismatches (disagreement, in orange) to question pairs on (a) completed task status, (b) partner knowledge, and (c) current task status asked at different relative intervals during games. Red crosses show the ratio of matched answers out of the total questions, and the red line shows the $3^{rd}$ order polynomial fit to the crosses.}
    \label{fig:hist_results}
\end{figure*}

\vspace{3pt}
\noindent
\textbf{Inferring Player Beliefs of Partner Current Task.} 
Here, we predict player beliefs of their partner's immediate task, specifically in response to questions such as ``What do you think the other player is making right now?''. For this question, participant answers can be one of 21 choices—the number of total possible material types participants may create in a game.
Compared to the predictions aforementioned, this experiment is more constrained in time to the vicinity of the question prompt. Our results in Figure \ref{fig:exp_results3} show that LSTM-based models seem to outperform transformer-based models, though only marginally in the video-only setting, demonstrating that local context seems to play a more important role in this prediction. 

\subsection{Analysis of the Evolution of Inferred Belief States}

As we are interested in the evolution of an agent's belief as the game progresses, we further plot prediction matches of \textit{model-predicted} belief states over every 10\% interval of the game, similar to that of \textit{player}-belief matches shown prior in Figure~\ref{fig:hist_human_label}. Figure \ref{fig:hist_results} shows the breakdowns from the best performing configuration for each experiment. 
For predicting the status of a completed task (Figure \ref{fig:hist_results1}), we observe that, similarly to human performance, the percentage of matched answers remains relatively stable, though we do notice a slight decrease for predictions later in the game. On the experiment of predicting the other player's knowledge, we see a similar increase in the percentage of matched answers as the game progresses (Figure \ref{fig:hist_results2}) as in Figure \ref{fig:hist_human_label2}. For the experiment of predicting the other player's current task (Figure \ref{fig:hist_results3}),  our model does not match the observations on human performance: the percentage of matched answers stays low and relatively constant. This result demonstrates that it is difficult to predict what the other player is doing given only interaction discourse and visual perception. This prediction requires a better understanding of the progress of the task, which the agent, as construed here, is lacking. This also points to the utility of actively engaging in dialogue\textemdash for example, explicitly asking what a partner is doing\textemdash to have a better understanding of their current goal.

\section{Conclusion and Future Work}

In a real-world collaborative scenario with physical agents, humans and agents will inevitably have disparities in their abilities, knowledge, and understanding of the shared world. This work specifically stimulates these disparities in a virtual environment, introducing a new dataset and experimental framework that supports in-depth studies of theory of mind modeling for situated dialogue in collaborative tasks. Through a novel implementation of self-reported belief states during collaborative interactions, our dataset keeps track of partners' beliefs about the task at hand and of each other step-by-step and captures how their states of mind evolve\textemdash and, indeed, how their common ground evolves\textemdash as communication and interaction unfold. To our knowledge, this is the first dataset in the context of situated dialogue that provides this fine-grained information for mental modeling. Our initial analysis of this dataset generates several interesting findings that will inform the development of computational models for various problems—for instance, in tracking mental models and managing dialogue behaviors in collaborative agents. Our baseline results demonstrate the importance of interaction discourse and visual experience in a shared environment on predicting mutual belief states of the task at hand, and of a collaborative partner, to ensure common ground.



While we have built baseline computational models to better help in understanding human collaborative behaviors and several theory of mind tasks, we hope our work further facilitates improvements in areas like agent planning and decision-making, computational reasoning, multimodal dialogue generation, and to move towards fully autonomous agents that are able to engage with humans in collaborative activities, on human terms, both effectively and efficiently in a human world.

\section*{Acknowledgements}

This work was supported by the National Science Foundation under grant IIS-1949634. The authors would like to thank those who participated in our data collection and the anonymous reviewers for their valuable comments and suggestions.


\section*{Ethical Considerations}

An application was submitted to the University of Michigan's Institutional Review Board (IRB) prior to the start of the project which deemed it exempt. A total of 29 subjects took part in the data collection; no personally identifiable information was stored throughout our experimental sittings, and participants were provided with anonymous Minecraft accounts to access the game servers such that they did not use their own. We do not additionally control for any ethnicity or cultural aspects aside from the condition that the participant is to be an English speaker and has some experience with Minecraft. 

\bibliography{emnlp2021}

\begin{thebibliography}{33}
\expandafter\ifx\csname natexlab\endcsname\relax\def\natexlab#1{#1}\fi

\bibitem[{Akula et~al.(2021)Akula, Wang, Liu, Saba-Sadiya, Lu, Todorovic, Chai,
  and Zhu}]{akula2021cxtom}
Arjun~R. Akula, Keze Wang, Changsong Liu, Sari Saba-Sadiya, Hongjing Lu, Sinisa
  Todorovic, Joyce Chai, and Song-Chun Zhu. 2021.
\newblock Cx-tom: Counterfactual explanations with theory-of-mind for enhancing
  human trust in image recognition models.
\newblock \emph{iScience, CELLPress, arXiv:2109.01401}.

\bibitem[{Bisk et~al.(2018)Bisk, Shih, Choi, and Marcu}]{bisk2018learning}
Yonatan Bisk, Kevin Shih, Yejin Choi, and Daniel Marcu. 2018.
\newblock Learning interpretable spatial operations in a rich 3d blocks world.
\newblock In \emph{Proceedings of the AAAI Conference on Artificial
  Intelligence}, volume~32.

\bibitem[{Bortolaso et~al.(2019)Bortolaso, Bourdiol, and
  Graham}]{bortolaso2019enhancing}
Christophe Bortolaso, J{\'e}r{\'e}my Bourdiol, and TC~Nicholas Graham. 2019.
\newblock Enhancing communication and awareness in asymmetric games.
\newblock In \emph{Joint International Conference on Entertainment Computing
  and Serious Games}, pages 250--262. Springer.

\bibitem[{Chai et~al.(2018)Chai, Gao, She, Yang, Saba-Sadiya, and
  Xu}]{chai2018language}
Joyce Chai, Qiaozi Gao, Lanbo She, Shaohua Yang, Sari Saba-Sadiya, and Guangyue
  Xu. 2018.
\newblock Language to action: Towards interactive task learning with physical
  agents.
\newblock In \emph{IJCAI}, pages 2--9.

\bibitem[{Chai et~al.(2014)Chai, She, Fang, Ottarson, Littley, Liu, and
  Hanson}]{chai2014collaborative}
Joyce Chai, Lanbo She, Rui Fang, Spencer Ottarson, Cody Littley, Changsong Liu,
  and Kenneth Hanson. 2014.
\newblock Collaborative effort towards common ground in situated human-robot
  dialogue.
\newblock In \emph{2014 9th ACM/IEEE International Conference on Human-Robot
  Interaction (HRI)}, pages 33--40. IEEE.

\bibitem[{Chandrasekaran et~al.(2017)Chandrasekaran, Yadav, Chattopadhyay,
  Prabhu, and Parikh}]{chandrasekaran2017takes}
Arjun Chandrasekaran, Deshraj Yadav, Prithvijit Chattopadhyay, Viraj Prabhu,
  and Devi Parikh. 2017.
\newblock It takes two to tango: Towards theory of ai's mind.
\newblock \emph{arXiv preprint arXiv:1704.00717}.

\bibitem[{Chung et~al.(2014)Chung, Gulcehre, Cho, and
  Bengio}]{chung2014empirical}
Junyoung Chung, Caglar Gulcehre, KyungHyun Cho, and Yoshua Bengio. 2014.
\newblock Empirical evaluation of gated recurrent neural networks on sequence
  modeling.
\newblock \emph{arXiv preprint arXiv:1412.3555}.

\bibitem[{Clark(1996)}]{clark1996using}
Herbert~H Clark. 1996.
\newblock \emph{Using language}.
\newblock Cambridge university press.

\bibitem[{Devlin et~al.(2019)Devlin, Chang, Lee, and
  Toutanova}]{Devlin2019BERTPO}
J.~Devlin, Ming-Wei Chang, Kenton Lee, and Kristina Toutanova. 2019.
\newblock Bert: Pre-training of deep bidirectional transformers for language
  understanding.
\newblock In \emph{NAACL-HLT}.

\bibitem[{Eicher et~al.(2017)Eicher, Cunningham, Gonzales, and
  Goel}]{eicher2017toward}
Bobbie Eicher, Kathryn Cunningham, Sydni Peterson~Marissa Gonzales, and Ashok
  Goel. 2017.
\newblock Toward mutual theory of mind as a foundation for co-creation.
\newblock In \emph{International Conference on Computational Creativity,
  Co-Creation Workshop}.

\bibitem[{Hochreiter and Schmidhuber(1997)}]{hochreiter1997long}
Sepp Hochreiter and J{\"u}rgen Schmidhuber. 1997.
\newblock Long short-term memory.
\newblock \emph{Neural computation}, 9(8):1735--1780.

\bibitem[{Iwahashi et~al.(2009)Iwahashi, Taguchi, Sugiura, Funakoshi, and
  Nakano}]{iwahashi2009robots}
Naoto Iwahashi, Ryo Taguchi, Komei Sugiura, Kotaro Funakoshi, and Mikio Nakano.
  2009.
\newblock Robots that learn to converse: Developmental approach to situated
  language processing.
\newblock In \emph{Proceedings of International Symposium on Speech and
  Language Processing}, pages 532--537.

\bibitem[{Jara-Ettinger(2019)}]{jara2019theory}
Julian Jara-Ettinger. 2019.
\newblock Theory of mind as inverse reinforcement learning.
\newblock \emph{Current Opinion in Behavioral Sciences}, 29:105--110.

\bibitem[{Jayannavar et~al.(2020)Jayannavar, Narayan-Chen, and
  Hockenmaier}]{jayannavar2020learning}
Prashant Jayannavar, Anjali Narayan-Chen, and Julia Hockenmaier. 2020.
\newblock Learning to execute instructions in a minecraft dialogue.
\newblock In \emph{Proceedings of the 58th Annual Meeting of the Association
  for Computational Linguistics}, pages 2589--2602.

\bibitem[{Laird et~al.(2017)Laird, Lebiere, and
  Rosenbloom}]{Laird_Lebiere_Rosenbloom_2017}
John~E Laird, Christian Lebiere, and Paul~S Rosenbloom. 2017.
\newblock A standard model of the mind: Toward a common computational framework
  across artificial intelligence, cognitive science, neuroscience, and
  robotics.
\newblock \emph{AI Magazine}, 38(4):13--26.

\bibitem[{Liu et~al.(2012)Liu, Fang, and Chai}]{liu12sigdial}
Changsong Liu, Rui Fang, and Joyce Chai. 2012.
\newblock Towards mediating shared perceptual basis in situated dialogue.
\newblock In \emph{Proceedings of the 13th Annual Meeting of the Special
  Interest Group on Discourse and Dialogue}, pages 140--149, Seoul, South
  Korea.

\bibitem[{Liu et~al.(2013)Liu, Fang, She, and Chai}]{liu13}
Changsong Liu, Rui Fang, Lanbo She, and Joyce Chai. 2013.
\newblock Modeling collaborative referring for situated referential grounding.
\newblock In \emph{Proceedings of the SIGDIAL 2013 Conference}, pages 78--86.

\bibitem[{McGuire et~al.(2002)McGuire, Fritsch, Steil, Rothling, Fink,
  Wachsmuth, Sagerer, and Ritter}]{mcguire2002multi}
Patrick McGuire, Jannik Fritsch, Jochen~J Steil, F~Rothling, Gernot~A Fink,
  Sven Wachsmuth, Gerhard Sagerer, and Helge Ritter. 2002.
\newblock Multi-modal human-machine communication for instructing robot
  grasping tasks.
\newblock In \emph{IEEE/RSJ International Conference on Intelligent Robots and
  Systems}, volume~2, pages 1082--1088. IEEE.

\bibitem[{Narayan-Chen et~al.(2019)Narayan-Chen, Jayannavar, and
  Hockenmaier}]{narayan2019collaborative}
Anjali Narayan-Chen, Prashant Jayannavar, and Julia Hockenmaier. 2019.
\newblock Collaborative dialogue in minecraft.
\newblock In \emph{Proceedings of the 57th Annual Meeting of the Association
  for Computational Linguistics}, pages 5405--5415.

\bibitem[{Popat and Palmer(2005)}]{popat2005creating}
Sita Popat and Scott Palmer. 2005.
\newblock Creating common ground: dialogues between performance and digital
  technologies.
\newblock \emph{International Journal of Performance Arts \& Digital Media},
  1(1).

\bibitem[{Powers et~al.(2005)Powers, Kramer, Lim, Kuo, Lee, and
  Kiesler}]{powers2005common}
Aaron Powers, Adam Kramer, Shirlene Lim, Jean Kuo, SL~Lee, and Sara Kiesler.
  2005.
\newblock Common ground in dialogue with a gendered humanoid robot.
\newblock \emph{Proceedings of RO-MAN 2005}.

\bibitem[{Premack and Woodruff(1978)}]{premack1978does}
David Premack and Guy Woodruff. 1978.
\newblock Does the chimpanzee have a theory of mind?
\newblock \emph{Behavioral and brain sciences}, 1(4):515--526.

\bibitem[{Qiu et~al.(2021)Qiu, Zhao, Liang, Lu, Shi, Yu, and
  Zhu}]{qiu2021towards}
Liang Qiu, Yizhou Zhao, Yuan Liang, Pan Lu, Weiyan Shi, Zhou Yu, and Song-Chun
  Zhu. 2021.
\newblock Towards socially intelligent agents with mental state transition and
  human utility.
\newblock \emph{arXiv preprint arXiv:2103.07011}.

\bibitem[{Rabinowitz et~al.(2018)Rabinowitz, Perbet, Song, Zhang, Eslami, and
  Botvinick}]{rabinowitz2018machine}
Neil Rabinowitz, Frank Perbet, Francis Song, Chiyuan Zhang, SM~Ali Eslami, and
  Matthew Botvinick. 2018.
\newblock Machine theory of mind.
\newblock In \emph{International conference on machine learning}, pages
  4218--4227. PMLR.

\bibitem[{Roman et~al.(2020)Roman, Bisk, Thomason, Celikyilmaz, and
  Gao}]{roman2020rmm}
Homero~Roman Roman, Yonatan Bisk, Jesse Thomason, Asli Celikyilmaz, and
  Jianfeng Gao. 2020.
\newblock Rmm: A recursive mental model for dialog navigation.
\newblock In \emph{Proceedings of the 2020 Conference on Empirical Methods in
  Natural Language Processing: Findings}, pages 1732--1745.

\bibitem[{Suhr et~al.(2019)Suhr, Yan, Schluger, Yu, Khader, Mouallem, Zhang,
  and Artzi}]{suhr2019executing}
Alane Suhr, Claudia Yan, Jack Schluger, Stanley Yu, Hadi Khader, Marwa
  Mouallem, Iris Zhang, and Yoav Artzi. 2019.
\newblock Executing instructions in situated collaborative interactions.
\newblock In \emph{Proceedings of the 2019 Conference on Empirical Methods in
  Natural Language Processing and the 9th International Joint Conference on
  Natural Language Processing (EMNLP-IJCNLP)}, pages 2119--2130.

\bibitem[{Thomason et~al.(2020)Thomason, Padmakumar, Sinapov, Walker, Jiang,
  Yedidsion, Hart, Stone, and Mooney}]{thomason2020jointly}
Jesse Thomason, Aishwarya Padmakumar, Jivko Sinapov, Nick Walker, Yuqian Jiang,
  Harel Yedidsion, Justin Hart, Peter Stone, and Raymond Mooney. 2020.
\newblock Jointly improving parsing and perception for natural language
  commands through human-robot dialog.
\newblock volume~67, pages 327--374.

\bibitem[{Udagawa and Aizawa(2019)}]{udagawa2019natural}
Takuma Udagawa and Akiko Aizawa. 2019.
\newblock A natural language corpus of common grounding under continuous and
  partially-observable context.
\newblock In \emph{Proceedings of the AAAI Conference on Artificial
  Intelligence}, volume~33, pages 7120--7127.

\bibitem[{Udagawa and Aizawa(2020)}]{udagawa2020annotated}
Takuma Udagawa and Akiko Aizawa. 2020.
\newblock An annotated corpus of reference resolution for interpreting common
  grounding.
\newblock In \emph{Proceedings of the AAAI Conference on Artificial
  Intelligence}, volume~34, pages 9081--9089.

\bibitem[{Vaswani et~al.(2017)Vaswani, Shazeer, Parmar, Uszkoreit, Jones,
  Gomez, Kaiser, and Polosukhin}]{vaswani2017attention}
Ashish Vaswani, Noam Shazeer, Niki Parmar, Jakob Uszkoreit, Llion Jones,
  Aidan~N Gomez, Lukasz Kaiser, and Illia Polosukhin. 2017.
\newblock Attention is all you need.
\newblock \emph{arXiv preprint arXiv:1706.03762}.

\bibitem[{Wang et~al.(2021)Wang, Saha, Gregori, Joyner, and
  Goel}]{wangCHI21tom}
Qiaosi Wang, Koustuv Saha, Eric Gregori, David Joyner, and Ashok Goel. 2021.
\newblock Towards mutual theory of mind in human-ai interaction: How language
  reflects what students perceive about a virtual teaching assistant.
\newblock In \emph{Proceedings of the 2021 CHI Conference on Human Factors in
  Computing Systems}, pages 1--14.

\bibitem[{Winograd(1972)}]{winograd1972understanding}
Terry Winograd. 1972.
\newblock Understanding natural language.
\newblock \emph{Cognitive psychology}, 3(1):1--191.

\bibitem[{Zarrie{\ss} et~al.(2016)Zarrie{\ss}, Hough, Kennington,
  Manuvinakurike, DeVault, Fern{\'a}ndez, and Schlangen}]{zarriess2016pentoref}
Sina Zarrie{\ss}, Julian Hough, Casey Kennington, Ramesh Manuvinakurike, David
  DeVault, Raquel Fern{\'a}ndez, and David Schlangen. 2016.
\newblock Pentoref: A corpus of spoken references in task-oriented dialogues.
\newblock In \emph{Proceedings of the Tenth International Conference on
  Language Resources and Evaluation (LREC'16)}, pages 125--131.

\end{thebibliography}
\bibliographystyle{acl_natbib}

\appendix

\newpage

\section{Appendices}

\subsection{Detailed Prediction Results}

A detailed comparison of the F1 scores on the testing and validation sets may be seen in Table \ref{tab:detail}. Each experiment was run 10 times; training each model for all settings lasts roughly 35 minutes on average.

\begin{table*}[h!]
    \centering
    \resizebox{0.8\textwidth}{!}{
    \begin{tabular}{ccc|r|r}
Task & Model & Setting & Test F1 Score & Validation F1 Score \\
\hline
\multirow{6}{*}{Predicting Completed Task Status } & \multirow{3}{*}{LSTM} & D   &  0.451($\pm$0.016) & 0.511($\pm$0.007) \\
&    &  D+V & 0.456($\pm$0.016) & 0.516($\pm$0.009) \\
&   &    V  & 0.450($\pm$0.021) & 0.519($\pm$0.008) \\
&  \multirow{3}{*}{Transformer}  &  D   & 0.398($\pm$0.022) & 0.486($\pm$0.009) \\
&   &  D+V & 0.439($\pm$0.034) & 0.512($\pm$0.023) \\
&   &    V & 0.536($\pm$0.015) & 0.549($\pm$0.009) \\
\hline

\multirow{6}{*}{Predicting Other Player's Knowledge} &  \multirow{3}{*}{LSTM} & D   &  0.490($\pm$0.010) & 0.690($\pm$0.004) \\
&    &  D+V & 0.488($\pm$0.009) & 0.679($\pm$0.007) \\
&   &    V  & 0.484($\pm$0.006) & 0.675($\pm$0.008) \\
&  \multirow{3}{*}{Transformer}  &  D   & 0.464($\pm$0.029) & 0.670($\pm$0.015) \\
&   &  D+V & 0.483($\pm$0.011) & 0.673($\pm$0.009) \\
&   &    V & 0.491($\pm$0.015) & 0.687($\pm$0.007) \\
\hline

\multirow{6}{*}{Predicting Other Player's Current Task} &  \multirow{3}{*}{LSTM} & D   &  0.081($\pm$0.012) & 0.140($\pm$0.006) \\
&    &  D+V & 0.070($\pm$0.013) & 0.137($\pm$0.005) \\
&   &    V  & 0.085($\pm$0.013) & 0.140($\pm$0.007) \\
&  \multirow{3}{*}{Transformer} &  D   & 0.047($\pm$0.013) & 0.133($\pm$0.009) \\
&   &  D+V & 0.056($\pm$0.006) & 0.129($\pm$0.005) \\
&   &    V & 0.076($\pm$0.014) & 0.149($\pm$0.007) \\
    \end{tabular}
}
    \caption{Model F1 scores on predicting player belief states on test and validation; 99\% confidence intervals are provided in parentheses.}
    \label{tab:detail}
\end{table*}





\subsection{Model Description}

\subsubsection{Convolutional Neural Network} The parameters for the convolutional network used in visual processing were primarily constrained by GPU memory limitations; image frames of size $96 \times 96$ were input into a CNN consisting of four convolutional layers with kernel sizes of $3 \times 3, 5 \times 5, 5 \times 5,$ and $3 \times 3$, whereby the sizes of the intermediate inputs were $3, 8, 32, 128,$ and $512$, respectively. These parameters were chosen under the consideration that the blocks-world Minecraft video frames are not as rich in content as that of a real-world photo setting. A Dropout of $0.2$ was further used between layers, chosen after a parameter sweep in the range of $[0,0.5]$, which was done by picking kernel sizes between $3$ and $5$ and layers between $3$ and $6$, taking into account the aforementioned image input size.  

 
\subsubsection{Plan Processing} Recall that each player is provided a partial view of the complete plan. For processing, each plan is stored as a list of tuples, representing all materials present in the plan and their links with (1) the materials needed to create them and (2) the tools needed to break them\textemdash i.e. nodes (materials) are linked with their children (composite materials and tools) as in the graph representation. The goal material, the root node, is always first in the list. All subsequent nodes are added in a breath first fashion, except in cases whereby a node has already been added to the list (as cycles are allowed). Each material and tool is given a one-hot encoding; mined materials have their children represented as zero vectors, as no other material is needed to make them. Partial plans and their representations are generated from the complete plan by hiding the children of randomly selected nodes\textemdash excluding the goal and mines\textemdash to depict a lack in knowledge. For encoding, each tuple has the one-hot encodings of (1) the material itself, (2) its parent node, (3) its children nodes, and (4) its associated tool concatenated; the list of tuples are then input to a GRU \cite{chung2014empirical}, which takes in an input vector of size 81 and has a hidden state size of 32. In the tasks that involve predicting a player's mental state from the perspective of the other player, only the partial plan associated to the other player's point of view is used. 


\subsection{Example Player Interaction}

\begin{figure*}[h!]
    \centering
    \begin{subfigure}{\textwidth}
        \includegraphics[width=1.0\textwidth]{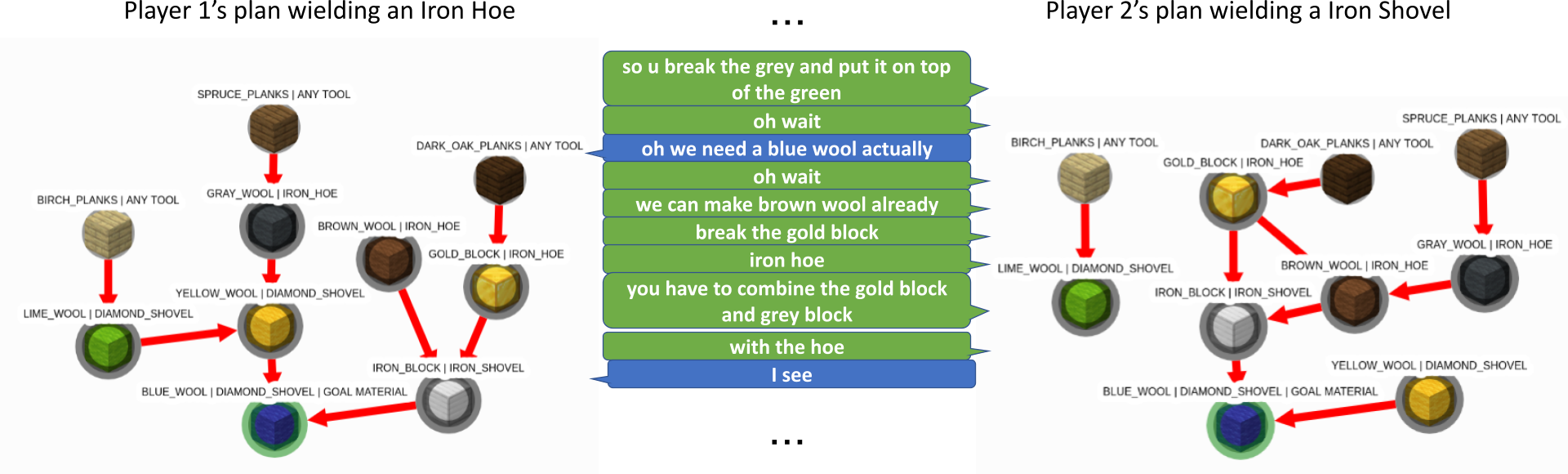}
        \caption{}
        \label{fig:example_dialogue1}
    \end{subfigure}
    \begin{subfigure}{\textwidth}
        \includegraphics[width=1.0\textwidth]{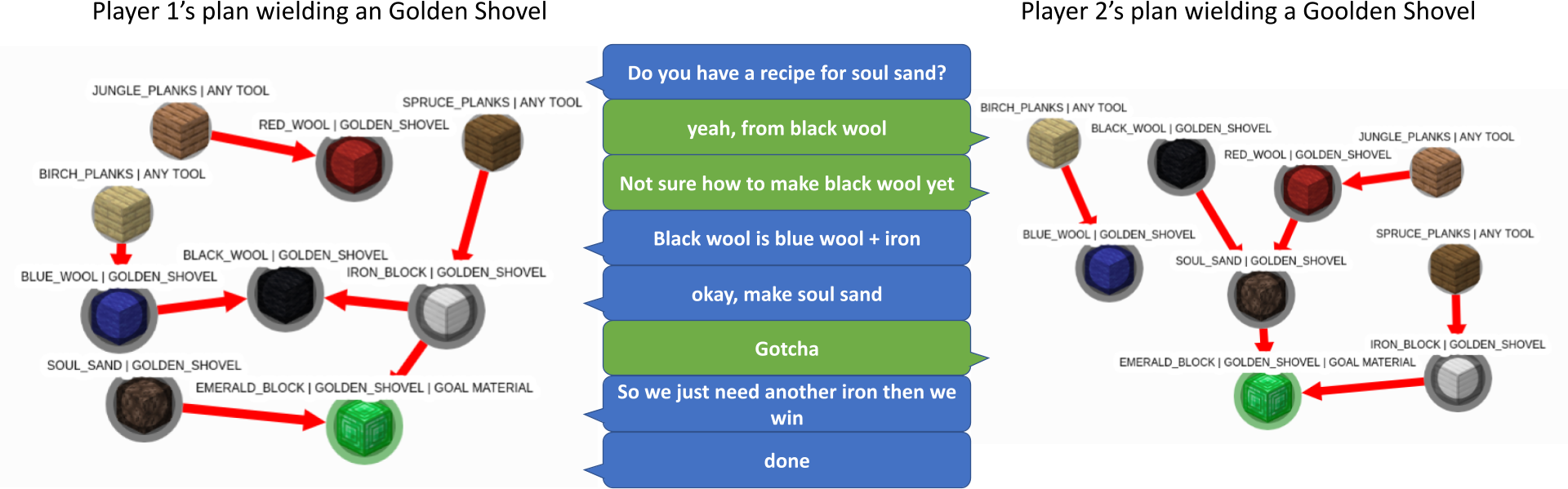}
        \caption{}
        \label{fig:example_dialogue2}
    \end{subfigure}
    \begin{subfigure}{\textwidth}
        \includegraphics[width=1.0\textwidth]{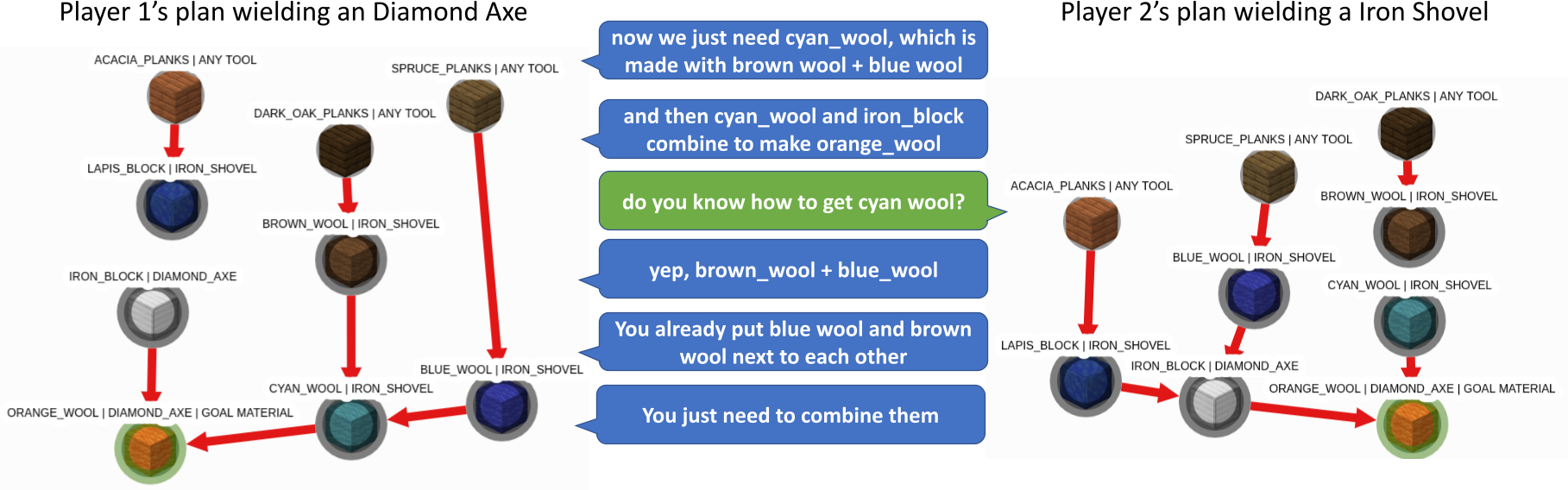}
        \caption{}
        \label{fig:example_dialogue3}
    \end{subfigure}
    
    \caption{Example dialogue exchanges, with the two players' partial plans also shown as context. 
    }
    \label{fig:example_dialogue}
\end{figure*}

Figure \ref{fig:example_dialogue1} shows a relatively verbose exchange of dialogue. Note that only a portion of the entire game's dialogue (which has 40 exchanges in total) is shown. Here, we observe that there is a clear self-assignment of leader and follower roles between the players: the leader explicitly states every step they think their partner needs to make, almost to the point of micro-managing. We also see an example of slight backtracking happening, where Player 2 realizes that they are further along in the plan than they initially thought.

In Figure \ref{fig:example_dialogue2}, we see an example of a fairly straightforward exchange of dialogue. Player 1 notices that they are not aware of the recipe for \texttt{Soul Sand}, which is needed to create the goal material, \texttt{Emerald Block}. They then inquire with their partner about it, who then states that they are unaware, instead, of how to make \texttt{Black Wool}, which is necessary for creating \texttt{Soul Sand}. Once the information is exchanged, the intermediate material is created promptly and both players then proceed to create their goal material.

Consider the dialogue exchange in Figure \ref{fig:example_dialogue3}. The two players are one step away from creating their goal material, \texttt{Orange Wool}. Player 1 points out that they require a block of \texttt{Cyan Wool}. Player 1 is pointing this out to Player 2 even though they cannot be sure Player 2 shares the knowledge as, in order to interact with the necessary materials, an \texttt{Iron Shovel}, which Player 1 does not possess, is required. From Player 2's perspective, while they are also aware that a block of \texttt{Cyan Wool} is required, they do not know how to make one as the arrows in their plan view are missing. As such, they inquire with their collaboration partner about the recipe. Player 1 then updates Player 2 on how to make \texttt{Cyan Wool} and also points out that one of the materials necessary was already created. This sample extract of their overall interaction is an example of grounding to the visual modality of their dialogue: our dataset provides much longer sequences of such interactions that are also causally dependant on one another. It is important to note here that the players are not assumed leader or follower roles; in this situation, the two participants coordinated entirely on their own and reached a consensus on who provides information and who is to execute the tasks. These roles switch throughout the game as their disparities in skills and knowledge change. 

These select dialogue exchanges showcase a small part of the diversity in possible interactions that happen in our experimental setup, whereby players are able to negotiate, decide, and execute their plans of action in a collaborative setting with relaxed constraints on player roles.

\end{document}